\documentclass[sigconf,anonymous=false]{acmart}

\AtBeginDocument{%
  \providecommand\BibTeX{{%
    \normalfont B\kern-0.5em{\scshape i\kern-0.25em b}\kern-0.8em\TeX}}}

\setcopyright{acmcopyright}
\copyrightyear{2022}
\acmYear{2022}
\setcopyright{rightsretained}
\acmConference[GECCO '22 Companion]{Genetic and Evolutionary Computation Conference Companion}{July 9--13, 2022}{Boston, MA, USA}
\acmBooktitle{Genetic and Evolutionary Computation Conference Companion (GECCO '22 Companion), July 9--13, 2022, Boston, MA, USA}
\acmDOI{10.1145/3520304.3529014}
\acmISBN{978-1-4503-9268-6/22/07}

\usepackage{cleveref}
\usepackage{todonotes}
\usepackage{interval}
\usepackage[per-mode = symbol]{siunitx}
\usepackage{booktabs, multirow}
\usepackage{caption, subcaption}

\usepackage{pgf}
\usepackage{tikz}
\usetikzlibrary{arrows.meta}

\usetikzlibrary{arrows}
\usetikzlibrary{calc}
\usetikzlibrary{shapes}
\usetikzlibrary{trees}
\usetikzlibrary{patterns}
\tikzset{>=stealth'}

\usetikzlibrary{automata}

\makeatletter
\renewcommand{\todo}[2][]{%
  \@todo[fancyline,caption={#2}, #1]{
    \begin{minipage}{\linewidth}
      \footnotesize #2
    \end{minipage}
  }%
}
\makeatother

\newcommand{\lcs}[1]{LCS#1}
\newcommand{\suprb}[1]{SupRB#1}

\begin{document}

\title{Separating Rule Discovery and Global Solution Composition in a Learning Classifier System
}

\author{Michael Heider}
\email{Michael.Heider@uni-a.de}
\orcid{0000-0003-3140-1993}
\affiliation{%
  \institution{Universit\"at Augsburg}
  \streetaddress{Am Technologiezentrum 8}
  \city{Augsburg}
  \country{Germany}
}
\author{Helena Stegherr}
\email{Helena.Stegherr@uni-a.de}
\orcid{0000-0001-7871-7309}
\affiliation{%
  \institution{Universit\"at Augsburg}
  \streetaddress{Am Technologiezentrum 8}
  \city{Augsburg}
  \country{Germany}
}
\author{Jonathan Wurth}
\email{Jonathan.Wurth@uni-a.de}
\orcid{0000-0002-5799-024X}
\affiliation{%
  \institution{Universit\"at Augsburg}
  \streetaddress{Am Technologiezentrum 8}
  \city{Augsburg}
  \country{Germany}
}
\author{Roman Sraj}
\email{Roman.Sraj@uni-a.de}
\affiliation{%
  \institution{Universit\"at Augsburg}
  \streetaddress{Am Technologiezentrum 8}
  \city{Augsburg}
  \country{Germany}
}
\author{J\"org H\"ahner}
\email{Joerg.Haehner@uni-a.de}
\orcid{0000-0003-0107-264X}
\affiliation{%
  \institution{Universit\"at Augsburg}
  \streetaddress{Am Technologiezentrum 8}
  \city{Augsburg}
  \country{Germany}
}

\renewcommand{\shortauthors}{Heider et al.}

\begin{abstract}
While utilization of digital agents to support crucial decision making is increasing, trust in suggestions made by these agents is hard to achieve.
However, it is essential to profit from their application, resulting in a need for explanations for both the decision making process and the model.
For many systems, such as common black-box models, achieving at least some explainability requires complex post-processing, while other systems profit from being, to a reasonable extent, inherently interpretable.
We propose a rule-based learning system specifically conceptualised and, thus, especially suited for these scenarios. 
Its models are inherently transparent and easily interpretable by design.
One key innovation of our system is that the rules' conditions and which rules compose a problem's solution are evolved separately. 
We utilise independent rule fitnesses which allows users to specifically tailor their model structure to fit the given requirements for explainability.
\end{abstract}




\maketitle

\section{Introduction}

With increasing automation and digitisation, interaction between humans and trained digital agents becomes more widespread.
Such socio-technical systems are for example encountered in smart factory settings.
Here, human stakeholders are dependent on recommendations made or decisions taken by an agent, e.g.\@ for (re-) configuring a machine.
However, at the moment complex learning tasks can rarely be solved perfectly---often because the available data for training is rather limited, e.g.\@ small sample, large imbalances.
This creates a distrust in the entire model among stakeholders, supposedly even if only edge cases were affected.
Hard to understand models even exacerbate this issue.
The cases of poor performance are rarely easily identifiable and even for good performance on test data, stakeholders often doubt the ability of models with a low transparency. 

A common approach to increase stakeholder trust in predictions is explaining the training and prediction processes themselves or the model in its entirety. 
With increasing model complexity, which is needed for difficult learning tasks, explaining the model or its predictions is less straightforward, leading to some types of models, e.g.\@ rule-based learners, being favoured for these situations, sometimes over better performing ones.
Learning Classifier Systems (\lcs{s}) are a family of rule-based learning algorithms that inherently allow application in the aforedescribed settings. \citep{Heider21}

\lcs{} models are composed of a finite set of if-then rules for which the conditions are optimized using a---typically evo\-lu\-tion\-ary---meta\-heu\-ri\-stic \citep{urbanowicz2009}.
Rules contain submodels of the problem that apply to certain areas of the feature space.
These submodels are comparatively simpler than models for the full problem, thus, increasing their comprehensibility by humans.
Most \lcs{s} follow the Michigan-style (a single set whose rules are adapted over time), featuring strong online learning capabilities and being employed to solve all major machine learning tasks.
However, these types of systems typically construct (and keep in their population) many more, sometimes suboptimal, rules than would be required to solve the problem at hand.
A common approach is, therefore, the reduction of the population to the essential rules after training has concluded, using so called \emph{compaction} techniques \citep{tan2013,liu2021b}.
The main other style \lcs{s} follow is the Pittsburgh-style.
Here, a population of sets of rules is evolved over time to solve learning tasks.
As a set of rules is assigned a combined fitness, rather than individual fitnesses for rules, optimal positioning/selection of rules is more difficult to achieve for the optimizer, especially as usually multiple changes to the set are performed per optimization step.
Suboptimal positioning does not necessarily substantially harm system performance. 
However, the importance of improving it increases when explanations for rule conditions or the training process are requested.
In general, the learning process is envisioned to create an ``accurate and maximally general'' \citep{urbanowicz2017}, ``maximally accurate and maximally general'' \citep{butz2004} or ``maximally general as well as accurate'' \citep{holland2000} set of rules.
Existing \lcs{} rarely specifically target explanations or transparency beyond the non formally specified requirement of generality, although they are still building interpretable models.

In this paper we present a new \lcs{} algorithm that is specifically designed to evolve both individual rules as well as the global problem solution (rule set), with performance as well as explainability considered during optimization.
To facilitate this we separate the optimization of rule conditions (find partitions of the feature space for which a submodel of the given type can be fit well) from optimizing a problem solution using these rules (cf. \Cref{suprb2}).
We name our system the \emph{Supervised Rule-based Learning System (SupRB)}, as it follows the same general goals as the Pittsburgh-style SupRB-1 \citep{heider2020a}.

\section{Related Work}
\label{related}

The most well known \lcs{s} are XCS and its derivatives \citep{urbanowicz2009}.
While XCS was originally designed for reinforcement learning tasks, it has (with some extensions) been applied in all of the three major learning settings \citep{urbanowicz2009}.
One of these extensions is the usage of interval-based matching functions rather than binary ones to operate in real-valued environments \cite{wilson2000a}.
To solve supervised function approximation tasks, XCS' constant predicted payoff was replaced with a linear function forming the original XCSF \citep{wilson2002}.
The linear model and the interval-based matching functions have later on been substituted with various more complex options \citep{bull2002b,lanzi2006}.
These are, however, sacrificing transparency for a stronger predictive performance.

Two approaches to reach \emph{explainability} and its related and relevant concepts of \emph{interpretability} and \emph{transparancy} and thus, ultimately, \emph{understandability} (in this paper we refer to those concepts under a broader umbrella of \emph{explainability} in the spirit of explainable artificial intelligence as a whole), must be distinguished \citep{barredoArrieta2020}:
By intentionally designing \emph{transparent models}, the model structure itself can be used for its comprehension and the interpretation of the decisions made. For other models, \emph{post-hoc methods} that operate through visualisation, transformation of complex black-box models into transparent models and other, often model-specific, techniques, need to be utilised.
Like other rule-based learning systems, \lcs{s} can, in general, be seen as transparent/interpretable by design.
They also relate to human behaviour naturally \citep{barredoArrieta2020}.
There are, however, some limitations that arise primarily through the encoding of variables, the size of the rule set and the complexity of individual rules. 

In \lcs{s} these limitations are typically controlled by design.
The variables themselves are problem dependent, so overall influence is limited, but using easy to understand matching functions that allow to follow the implications for decision boundaries in the feature space improves model transparency.
However, if the variable/feature itself is highly complex, human interpretation is always limited.
Rule complexity is likewise chosen by using a fitting submodel to balance users' transparency requirements with predictive power.
Additionally, human understanding can be improved post-hoc by employing a variety of different visualisation techniques for classifiers \citep{urbanowicz2012,liu2019,liu2021}.

In contrast to these generally applicable solutions, handling rule set size is approached differently depending on the \lcs{(-style)}.
Pittsburgh-style \lcs{s} can control set size directly via their fitness function.
For example, GAssist \citep{bacardit2004} can use the minimum description length in combination with accuracy to form a single objective fitness and additionally apply a penalty on individuals' fitnesses when the rule set size falls below a predefined threshold.
In Michigan-style systems, where fitness refers to a rule rather than a rule set and training benefits from larger than necessary populations, similar mechanisms are not available.
Instead, \emph{compaction} mechanisms have been designed \cite{wilson2002a,dixon2003}.
After training is completed, they remove redundant or incorrect rules from the population.
Ideally, the rule set size decreases without a negative effect on predictive power.
This has first been demonstrated \citep{wilson2001} on the Wisconsin Breast Cancer dataset and further advanced until the issue was considered solved by \citet{tan2013}.
Recently, \citet{liu2021b} have proposed new compaction techniques and demonstrated their improvements over existing methods on a variety of boolean benchmarking and three real world problems.

There are some hybrid rule-based learning systems which combine explicit Michigan- and Pittsburgh-style phases for improving explainability by reducing the number of rules \citep{ishibuchi2004,ishibuchi2005,chan2010,dutta2020}.
Most utilise the same evolutionary algorithm for both phases, often some multi-objective evolutionary algorithm to find a proper balance between the number of rules and the accuracy. 
Furthermore, the two phases can be applied subsequently \citep{chan2010}, nested \citep{ishibuchi2005} or cyclic, where both phases are executed several times \citep{dutta2020}.


\section{The Supervised Rule-based Learning System}
\label{suprb2}

The main idea of \suprb{} is to optimize rule conditions independently of other rules, discovering a diverse pool of well proportioned rules and then use another optimization process to select a good subset of all available rules to find good solutions to the learning problem.
By separating these optimization processes both can include multiple objectives to improve explainability of the \lcs{} model, while still targeting overall performance, e.g.\@ rules should encompass large feature spaces but be positioned to allow a well fitted submodel and solutions should be composed of only few rules while still minimizing prediction error.
A Python implementation of \suprb{} is available on GitHub\footnote{\url{https://github.com/heidmic/suprb} $\quad$ \url{https://doi.org/10.5281/zenodo.6460701}}.

\begin{figure}
\begin{tikzpicture}[remember picture,->,>=stealth',shorten >=1pt,auto,node distance=4cm,semithick,initial text=${n\leftarrow0}$]
  \tikzstyle{every state}=[fill=none,draw=black,text=black,align=center]

  \node[initial,state,minimum size=2.1cm] (D)                    {Discover\\ Rules};
  \node[state,minimum size=2.1cm]         (C) [below of=D]       {Compose\\ Solution\\ from Pool};
  \node[state,accepting,yshift=2cm,minimum size=0.1cm](T)[below of=C]		  {};

  \path (D) edge[loop above] node {$-$ / $n \leftarrow n+1$}	(D)
            edge[bend left]	node[align=right] {
            \begin{tabular}{p{1.3cm} @{\hspace{0.5\tabcolsep}} p{2.5cm}}
            $n=n_{RD}$ / & \raggedright put rules into pool, $n\leftarrow0$
            \end{tabular} }	(C)
		(C) edge[bend left] node[align=right] {$n=n_{SC}$ / $n\leftarrow0$}	  	(D)
			edge[loop right] node {$-$ / $n \leftarrow n+1$}		(C)
			edge			  node[align=center] {termination criterion / $-$}		(T);
\end{tikzpicture}
\caption{Rule discovery and solution composition phases in \suprb{}. \textmd{$n_{SC}$ denotes the number of steps the solution creating/composing optimizer should undertake, while $n_{RD}$ references the number of steps performed within rule discovery.}}
\label{statemachine}
\end{figure}
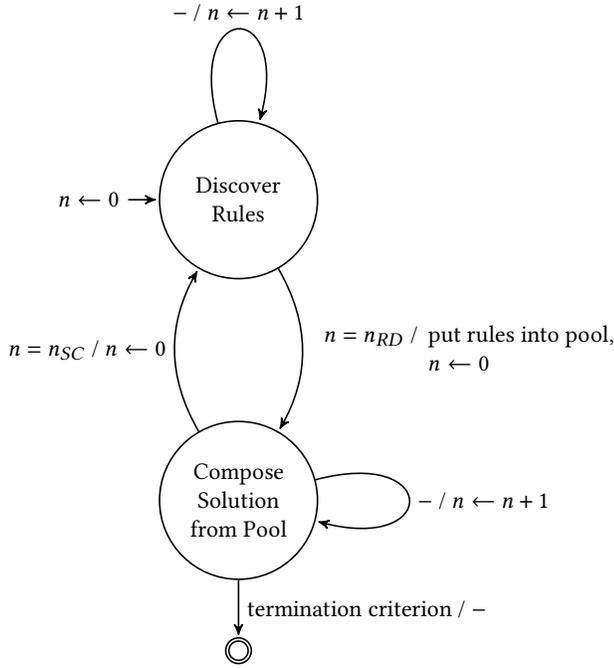

For unknown problems, it is hard to estimate how many rules will likely need to be discovered before a good subset can be selected.
Therefore, we alternate between phases of discovering new rules and combining rules from the pool of discovered and fitted rules.
The expectation is that we can find a good solution with fewer submodel fittings than with conservative estimates of needed rules, while still being able to find such a solution if the number of rules was underestimated.
Note that rules added to the pool remain unchanged and will not be removed throughout the training process.
Another advantage of alternating between phases is that we can use information from the solution composition phase to steer subsequent rule discoveries towards exploring regions where no or ill-placed rules are found.
The overall process is illustrated in the form of a statemachine in \Cref{statemachine}. 
The number of optimization steps performed within each phase can be varied, which can impact overall convergence time but does not affect solution strength. 

As deriving insights into decisions (cf. \Cref{related}) is a central aspect of \suprb{}, its model is deliberately kept as simple as possible:
\begin{enumerate}
\item Rules' conditions use an interval based matching: 
A rule $k$ applies for example $x$ iff $x_i \in [l_{k, i}, u_{k, i}] \forall i$ with $l$ being the lower and $u$ the upper bounds.
\item Rules' submodels $f_k(x)$ are linear.
They are fit using linear least squares with a l2-norm regularization (Ridge Regression) on the subsample matched by the respective rule.
\item When mixing multiple rules to make a prediction, a rule's experience (the number of examples matched during training and therefore included in fitting the submodel) and in-sample error are used in a weighted sum.
\end{enumerate}

\subsection{Rule Discovery}
To discover a new rule for the pool we use an evolution strategy (ES). 
Note that, in contrast to the hybrid systems described at the end of \Cref{related}, this rule discovery approach can not be considered a Michigan-style phase, especially as new rules are evolved one at a time.
Its initial individual is placed around a randomly selected training example, prioritizing examples that have a high in-sample error in the current (intermediate) global solution.
Then we use a mutation operator without adaptation that samples a halfnormal distribution twice and moves the upper and lower bound further from the center, according to the respective values, to create $\lambda$ children.
From these, we replace the parent with the individual that has the highest fitness based on its in-sample error and the matched feature space volume.
Specifically, the fitness is calculated as 
\begin{equation}
  \label{eq:fitness}
  F(o_1, o_2) = \frac{(1 + \alpha^2) \cdot o_1 \cdot o_2}{\alpha^2 \cdot o_1 + o_2} \,,
\end{equation}
with
\begin{equation}
  \label{eq:pseudo-accuracy}
  o_1 = \text{PACC} = \exp(-\text{MSE} \cdot \beta) \,,
\end{equation}
and
\begin{equation}
  o_2 = V = \prod_{i}{\frac{u_{i} - l_{i}}{\min_{x \in \mathcal{X}}{x_i} - \max_{x \in \mathcal{X}}{x_i}}} \,.
\end{equation} 
Its base form (cf.\@ \cref{eq:fitness}) was adapted from~\cite{wu2019}, where it was used in a feature selection context, similarly combining two objectives.
The Pseudo-Accuracy (PACC) squashes the Mean Squared Error (MSE) of a rule's prediction into a $(0, 1]$ range, while the volume share $V \in [0, 1]$ of its bounds is used as a generality measure.
The parameter $\beta = 2$ controls the slope of the PACC and $\alpha$ weighs the importance of $o_1$ against $o_2$.
Maximizing both objectives hence corresponds to generating rules that have minimal error and are maximally general.
A special form of plus-selection is used in the ES, which simultaneously controls the number of iterations: for every iteration, the best of $\lambda$ children is saved as an elitist and compared with all elitists from previous iterations. 
If the elitist from $\delta$ iterations before is better than all subsequent elitists, the optimization process is stopped and this specific elitist is added to the pool.
This procedure to discover a new rule for the pool is performed multiple times before this phase ends.
As this optimizer is not meant to find a globally optimal rule but rather fill a multitude of niches with optimally placed rules, independent evolution is advantageous.

\subsection{Solution Creation}
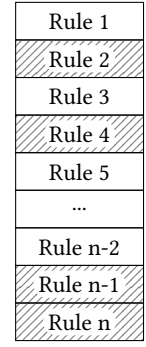
\begin{figure}
\begin{tikzpicture}[every node/.style={draw=none,rectangle,minimum width=1.7cm,minimum height= 0.5cm,align=center}]

\draw node[draw=black,yshift=0.5cm] (cla) {Rule 1};
\draw node[draw=black,pattern=north east lines,pattern color=black!50,below of=cla,yshift=0.5cm] (clb) {Rule 2};
\draw node[draw=black,below of=clb,yshift=0.5cm] (clc) {Rule 3};
\draw node[draw=black,pattern=north east lines,pattern color=black!50,below of=clc,yshift=0.5cm] (cld) {Rule 4};
\draw node[draw=black,below of=cld,yshift=0.5cm] (cle) {Rule 5};
\draw node[draw=black,below of=cle,yshift=0.5cm] (clf) {...};
\draw node[draw=black,below of=clf,yshift=0.5cm] (clw) {Rule n-2};
\draw node[draw=black,pattern=north east lines,pattern color=black!50,below of=clw,yshift=0.5cm] (clx) {Rule n-1};
\draw node[draw=black,pattern=north east lines,pattern color=black!50,below of=clx,yshift=0.5cm] (cly) {Rule n};

\draw node[fill=white,rounded corners=5pt,below of=cla,minimum width=0.5cm,minimum height= 0.2cm,yshift=0.5cm,inner xsep=2pt, inner ysep=2pt] (clb2) {Rule 2};
\draw node[fill=white,rounded corners=5pt,below of=clc,minimum width=0.5cm,minimum height= 0.2cm,yshift=0.5cm,inner xsep=2pt, inner ysep=2pt] (cld2) {Rule 4};
\draw node[fill=white,rounded corners=5pt,below of=clw,minimum width=0.5cm,minimum height= 0.2cm,yshift=0.5cm,inner xsep=2pt, inner ysep=2pt] (clx2) {Rule n-1};
\draw node[fill=white,rounded corners=5pt,below of=clx,minimum width=0.5cm,minimum height= 0.2cm,yshift=0.5cm,inner xsep=2pt, inner ysep=2pt] (clx3) {Rule n};

\end{tikzpicture}
\caption{Example global problem solution for a pool of size n. Selected rules are highlighted. In binary notation (on which the optimizer operates) this individual corresponds to $0 1 0 1 0 \dots 0 1 1$.}
\label{solution}
\end{figure}

After new rules have been discovered, a genetic algorithm (GA) selects rules from the pool to form a new solution candidate (a set of rules).
Solution candidates are represented as bit strings, signalling whether a rule from the pool is part of the candidate (cf.\@ \Cref{solution}).
The GA is configured to use tournament selection and combine two candidate solutions using $n$-point crossover with a crossover probability. 
Afterwards, each individual bit of the children is flipped with a probability given by the mutation rate.  
The candidate fitness is similarly based on \cref{eq:fitness}, using the candidate's in-sample MSE and \emph{complexity}, i.e. the number of rules selected, as first and second objective, respectively.
A certain number of elitist solutions from the previous population is additionally copied to the new population without modification.
Note that individuals in the GA always form a subset of the pool. 
Rules that are not part of the pool can not be part of a solution candidate and rules remain unchanged by the GA's operations, in contrast to typical Pittsburgh-style systems.

\section{Conclusion}

We presented a new Learning Classifier System (LCS) that separates the process of rule discovery from the composition of rules to form problem solutions.
This system performs supervised batch-learning and is therefore called the Supervised Rule-based Learning System (\suprb{}).
It utilizes a population-based optimizer (genetic algorithm) whose individuals transcribe which rules from the pool of discovered and locally optimized rules are part of a solution.
In contrast to many Pittsburgh-style approaches, which also evolve populations of rule sets, the rules from the pool are always and automatically available to all individuals.
Optimization of individuals combines a fitness pressure for low errors and low complexities (number of rules).
To fill the pool with rules, we utilized a simplistic ES ($\mu=1$) that optimizes towards low in-sample errors and high volumes of matched feature space.
Note that in contrast to other similar systems, the rule discovery is not done in a Michigan-style phase. 
Rules are added sequentially in separated evolutionary processes and fitnesses are independent from each other.

The primary motivation for our new approach at creating LCS models was to achieve a greater and more direct control over rule set sizes and matching functions, and thus the overall model structure. 
Finding a good model structure is also known as the model selection problem.
Ultimately, this leads to more interpretable models that make providing explanations for both the model itself, as well as its predictions, easier. 


\bibliographystyle{ACM-Reference-Format}
\bibliography{References}

\end{document}